\documentclass[10pt,twocolumn,letterpaper]{article}

\usepackage{iccp}
\usepackage{times}
\usepackage[utf8]{inputenc} 
\usepackage[T1]{fontenc}    
\usepackage{hyperref}       
\usepackage{url}            
\usepackage{booktabs}       
\usepackage{amsfonts}       
\usepackage{nicefrac}       
\usepackage{microtype}      
\usepackage{graphicx}
\usepackage{amsmath}
\usepackage{amssymb}
\usepackage{authblk}
\usepackage{abstract}
\usepackage{float}

\iccpfinalcopy 

\def\iccpPaperID{6} 

\ificcpfinal\fi
\begin{document}

\title{A Learning-based Framework for Hybrid Depth-from-Defocus and Stereo Matching}

\newcommand\CoAuthorMark{\footnotemark[\arabic{footnote}]}
\newcommand*\samethanks[1][\value{footnote}]{\footnotemark[#1]}
\renewcommand\Authands{ and }

\author[1]{Zhang Chen\thanks{These authors contribute to the work equally.}}
\author[2]{Xinqing Guo\samethanks}
\author[1]{Siyuan Li}
\author[1]{Xuan Cao}
\author[1]{Jingyi Yu}
\affil[1]{ShanghaiTech University, Shanghai, China. \texttt{\{chenzhang, lisy1, caoxuan, yujingyi\}@shanghaitech.edu.cn}}
\affil[2]{University of Delaware, Newark, DE, USA. \texttt{xinqing@udel.edu}}

\maketitle
\thispagestyle{empty}

\begin{abstract}
Depth from defocus (DfD) and stereo matching are two most studied passive depth sensing schemes. The techniques are essentially complementary: DfD can robustly handle repetitive textures that are problematic for stereo matching whereas stereo matching is insensitive to defocus blurs and can handle large depth range. In this paper, we present a unified learning-based technique to conduct hybrid DfD and stereo matching. Our input is image triplets: a stereo pair and a defocused image of one of the stereo views. We first apply depth-guided light field rendering to construct a comprehensive training dataset for such hybrid sensing setups. Next, we adopt the hourglass network architecture to separately conduct depth inference from DfD and stereo. Finally, we exploit different connection methods between the two separate networks for integrating them into a unified solution to produce high fidelity 3D disparity maps. Comprehensive experiments on real and synthetic data show that our new learning-based hybrid 3D sensing technique can significantly improve accuracy and robustness in 3D reconstruction.
\end{abstract}

\section{Introduction}

Acquiring 3D geometry of the scene is a key task in computer vision. Applications are numerous, from classical object reconstruction and scene understanding to the more recent visual SLAM and autonomous driving. Existing approaches can be generally categorized into active or passive 3D sensing. Active sensing techniques such as LIDAR and structured light offer depth map in real time but require complex and expensive imaging hardware. Alternative passive scanning systems are typically more cost-effective and can conduct non-intrusive depth measurements but maintaining its robustness and reliability remains challenging.

Stereo matching and depth from defocus (DfD) are the two best-known passive depth sensing techniques. Stereo recovers depth by utilizing parallaxes of feature points between views. At its core is correspondences matching between feature points and patching the gaps by imposing specific priors, e.g., induced by the Markov Random Field. DfD, in contrast, infers depth by analyzing blur variations at same pixel captured with different focus settings (focal depth, apertures, etc). Neither technique, however, is perfect on its own: stereo suffers from ambiguities caused by repetitive texture patterns and fails on edges lying along epipolar lines whereas DfD is inherently limited by the aperture size of the optical system.

It is important to note that DfD and stereo are complementary to each other: stereo provides accurate depth estimation even for distant objects whereas DfD can reliably handle repetitive texture patterns. In computational imaging, a number of hybrid sensors have been designed to combine the benefits of the two. In this paper, we seek to leverage deep learning techniques to infer depths in such hybrid DfD and stereo setups. Recent advances in neural network have revolutionized both high-level and low-level vision by learning a non-linear mapping between the input and output. Yet most existing solutions have exploited only stereo cues \cite{luo16, zagoruyko15, zbontar15} and very little work addresses using deep learning for hybrid stereo and DfD or even DfD alone, mainly due to the lack of a fully annotated DfD dataset.

In our setup, we adopt a three images setting: an all-focus stereo pair and a defocused image of one of the stereo views, the left in our case. We have physically constructed such a hybrid sensor by using Lytro Illum camera. We first generate a comprehensive training dataset for such an imaging setup. Our dataset is based on FlyingThings3D from \cite{mayer16}, which contains stereo color pairs and ground truth disparity maps. We then apply occlusion-aware light field rendering\cite{yang16} to synthesize the defocused image. Next, we adopt the hourglass network \cite{newell16} architecture to extract depth from stereo and defocus respectively. Hourglass network features a multi-scale architecture that consolidates both local and global contextures to output per-pixel depth. We use stacked hourglass network to repeat the bottom-up, top-down depth inferences, allowing for refinement of the initial estimates. Finally, we exploit different connection methods between the two separate networks for integrating them into a unified solution to produce high fidelity 3D depth maps. Comprehensive experiments on real and synthetic data show that our new learning-based hybrid 3D sensing technique can significantly improve accuracy and robustness in 3D reconstruction.

\subsection{Related Work}

\noindent\textbf{Learning based Stereo}
Stereo matching is probably one of the most studied problems in computer vision. We refer the readers to the comprehensive survey \cite{scharstein02, brown03}. Here we only discuss the most relevant works. Our work is motivated by recent advances in deep neural network. One stream focuses on learning the patch matching function. The seminal work by {\v Z}bontar and LeCun ~\cite{zbontar2016stereo} leveraged convolutional neural network (CNN) to predict the matching cost of image patches, then enforced smoothness constraints to refine depth estimation. \cite{zagoruyko15} investigated multiple network architectures to learn a general similarity function for wide baseline stereo. Han \emph{et al.} \cite{han2015matchnet} described a unified approach that includes both feature representation and feature comparison functions. Luo \emph{et al.} \cite{luo16} used a product layer to facilitate the matching process, and formulate the depth estimation as a multi-class classification problem. Other network architectures \cite{chen2015deep, liu2016euclidean, park2016look} have also been proposed to serve a similar purpose.

Another stream of studies exploits end-to-end learning approach. Mayer \emph{et al.} \cite{mayer16} proposed a multi-scale network with contractive part and expanding part for real-time disparity prediction. They also generated three synthetic datasets for disparity, optical flow and scene flow estimation. Kn{\"o}belreiter \emph{et al.}~\cite{knobelreiter2016end} presented a hybrid CNN+CRF model. They first utilized CNNs for computing unary and pairwise cost, then feed the costs into CRF for optimization. The hybrid model is trained in an end-to-end fashion. In this paper, we employ end-to-end learning approach for depth inference due to its efficiency and compactness.

\noindent\textbf{Depth from Defocus}
The amount of blur at each pixel carries information about object's distance, which could benefit numerous applications, such as saliency detection \cite{li2014saliency, li2015weighted}. To recover scene geometry, earlier DfD techniques \cite{subbarao94, rajagopalan97, watanabe98} rely on images captured with different focus settings (moving the objects, the lense or the sensor, changing the aperture size, etc). More recently, Favaro and Soatto \cite{favaro07} formulated the DfD problem as a forward diffusion process where the amount of diffusion depends on the depth of the scene. \cite{levin07, zhou10} recovered scene depth and all-focused image from images captured by a camera with binary coded aperture. Based on a per-pixel linear constraint from image derivatives, Alexander \emph{et al.} \cite{alexander16} introduced a monocular computational sensor to simultaneously recover depth and motion of the scene.

Varying the size of the aperture \cite{pentland87, ens91, surya93, bove93} has also been extensively investigated. This approach will not change the distance between the lens and sensor, thus avoiding the magnification effects. Our DfD setting uses a defocused and all-focused image pair as input, which can be viewed as a special case of the varying aperture. To tackle the task of DfD, we utilize a multi-scale CNN architecture. Different from conventional hand-crafted features and engineered cost functions, our data-driven approach is capable of learning more discriminative features from the defocus image and inferring the depth at a fraction of the computational cost.

\noindent\textbf{Hybrid Stereo and DfD Sensing}
In the computational imaging community, there has been a handful of works that aim to combine stereo and DfD. Early approaches \cite{klarquist95, subbarao97} use a coarse estimation from DfD to reduce the search space of correspondence matching in stereo. Rajagopalan \emph{et al.} \cite{Rajagopalan04} used a defocused stereo pair to recover depth and restore the all-focus image. Recently, Tao \emph{et al.} \cite{tao13} analyzed the variances of the epipolar plane image (EPI) to infer depth: the horizontal variance after vertical integration of the EPI encodes the defocus cue, while vertical variance represents the disparity cue. Both cues are then jointly optimized in a MRF framework. Takeda \emph{et al.} \cite{takeda13} analyzed the relationship between point spread function and binocular disparity in the frequency domain, and jointly resolved the depth and deblurred the image. Wang \emph{et al.} \cite{wang16} presented a hybrid camera system that is composed of two calibrated auxiliary cameras and an uncalibrated main camera. The calibrated cameras were used to infer depth and the main camera provides DfD cues for boundary refinement. Our approach instead leverages the neural network to combine DfD and stereo estimations. To our knowledge, this is the first approach that employs deep learning for stereo and DfD fusion.

\section{Training Data}
\label{sec:dataGeneration}

\begin{figure*}[t]
\begin{center}
   \includegraphics[width=0.99\linewidth]{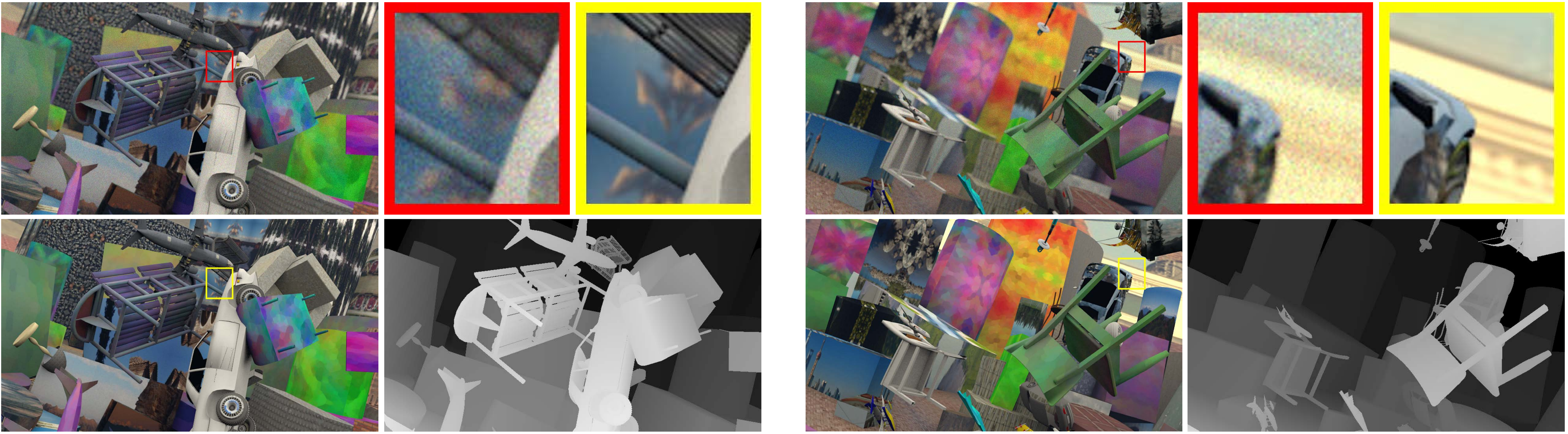}
\end{center}
\vspace{-8pt}
   \caption{The top row shows the generated defocused image by using \emph{Virtual DSLR} technique. The bottom row shows the ground truth color and depth images. We add Poisson noise to training data, a critical step for handling real scenes. The close-up views compare the defocused image with noise added and the original all-focus image.}
\label{fig:DfD_dataset}
\end{figure*}

The key to any successful learning based depth inference scheme is a plausible training dataset. Numerous datasets have been proposed for stereo matching but very few are readily available for defocus based depth inference schemes. To address the issue, we set out to create a comprehensive DfD dataset. Our DfD dataset is based on FlyingThing3D \cite{mayer16}, a synthetic dataset consisting of everyday objects randomly placed in the space. When generating the dataset, \cite{mayer16} separates the 3D models and textures into disjointed training and testing parts. In total there are 25,000 stereo images with ground truth disparities. In our dataset, we only select stereo frames whose largest disparity is less than 100 pixels to avoid objects appearing in one image but not in the other.

The synthesized color images in FlyingThings3D are all-focus images. To simulate defocused images, we adopt the \emph{Virtual DSLR} approach from \cite{yang16}. \emph{Virtual DSLR} uses color and disparity image pair as input and outputs defocused image with quality comparable to those captured by expensive DSLR. The algorithm resembles refocusing technique in light field rendering without requiring the actual creation of the light field, thus reducing both memory and computational cost. Further, the \emph{Virtual DSLR} takes special care of occlusion boundaries, to avoid color bleeding and discontinuity commonly observed on brute-force blur-based defocus synthesis.

For the scope of this paper, we assume circular apertures, although more complex ones can easily be synthesized, e.g., for coded-aperture setups. To emulate different focus settings of the camera, we randomly set the focal plane, and select the size of the blur kernel in the range of $7\sim23$ pixels. Finally, we add Poisson noise to both defocused image and the stereo pair to simulate the noise contained in real images. We'd emphasize that the added noise is critical in real scene experiments, as will be discussed in \ref{sec:realExperiments}. Our final training dataset contains 750 training samples and 160 testing samples, with each sample containing one stereo pair and the defocused image of the left view. The resolution of the generated images is $960\times540$, the same as the ones in FlyingThings3D. Figure \ref{fig:DfD_dataset} shows two samples of our training set.

\section{DfD-Stereo Network Architecture}
\label{sec:networkArchitecture}
\begin{figure*}[t]
\begin{center}
   \includegraphics[width=1.0\linewidth]{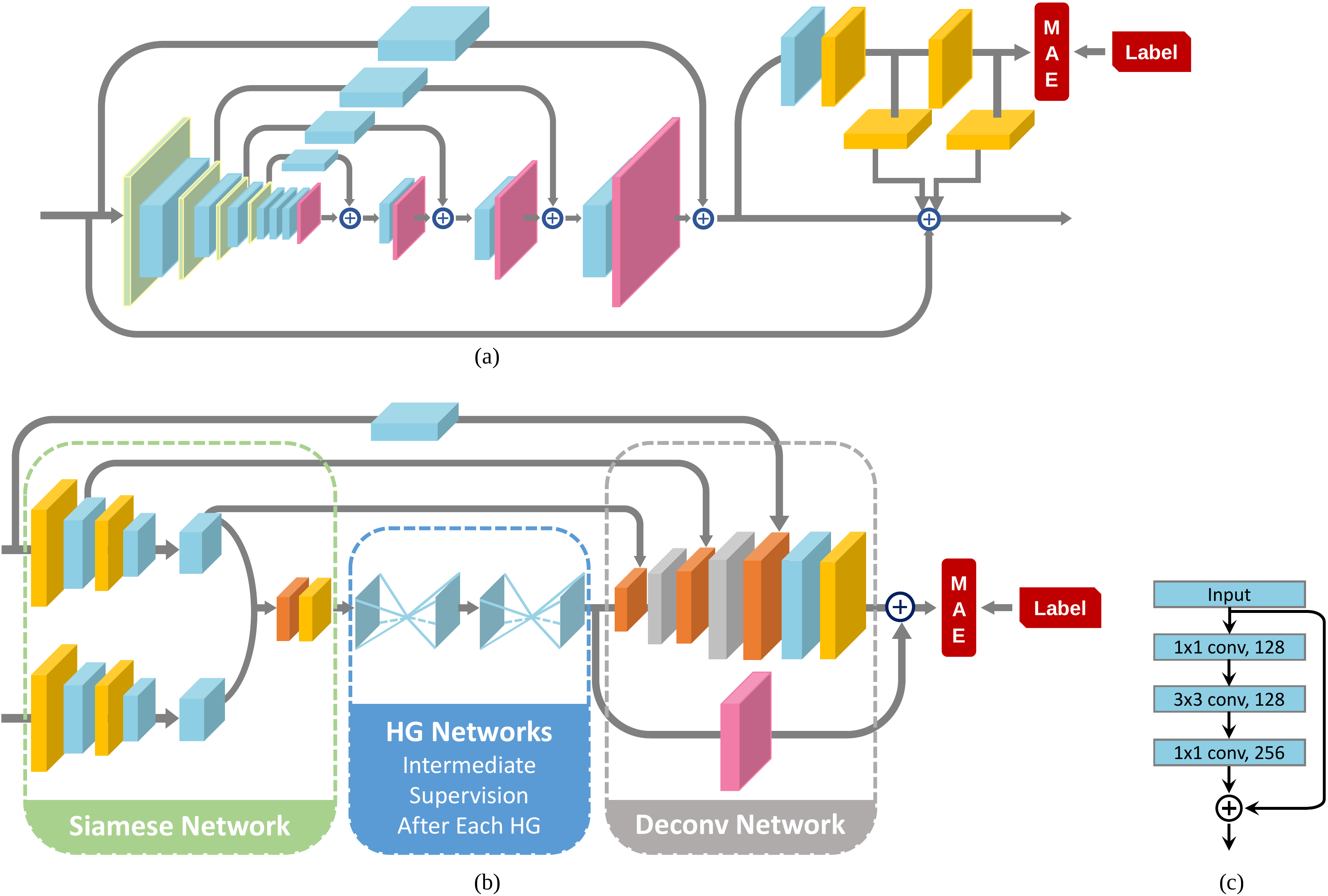}
\end{center}
\vspace{-8pt}
   \caption{(a) The hourglass network architecture consists of the max pooling layer (green), the nearest neighbor upsampling layer (pink), the residual module (blue), and convolution layer (yellow). The network includes intermediate supervision (red) to facilitate the training process. The loss function we use is mean absolute error (MAE). (b) The overall architecture of HG-DfD-Net and HG-Stereo-Net. The siamese network before the HG network aims to reduce the feature map size, while the deconvolution layers (gray) progressively recover the feature map to its original resolution. At each scale, the upsampled low resolution features are fused with high resolution features by using the concatenating layer (orange) (c) shows the detailed residual module.}
\label{fig:network_hourglass}
\end{figure*}

Depth inference requires integration of both fine- and large-scale structures. For DfD and stereo, the depth cues could be distributed at various scales in an image. For instance, textureless background requires the understanding of a large region, while objects with complex shapes need attentive evaluation of fine details. To capture the contextual information across different scales, a number of recent approaches adopt multi-scale networks and the corresponding solutions have shown plausible results \cite{eigen14, hui16}. In addition, recent studies \cite{simonyan14} have shown that a deep network with small kernels is very effective in image recognition tasks. In comparison to large kernels, multiple layers of small kernels maintain a large receptive field while reducing the number of parameters to avoid overfitting. Therefore, a general principle in designing our network is a deep multi-scale architecture with small convolutional kernels.

\subsection{Hourglass Network for DfD and Stereo}
Based on the observations above, we construct multi-scale networks that follow the hourglass (HG) architecture \cite{newell16} for both DfD and stereo. Figure \ref{fig:network_hourglass} illustrates the structure of our proposed network.

HG network features a contractive part and an expanding part with skip layers between them. The contractive part is composed of convolution layers for feature extraction, and max pooling layers for aggregating high-level information over large areas. Specifically, we perform several rounds of max pooling to dramatically reduce the resolution, allowing smaller convolutional filters to be applied to extract features that span across the entire space of image. The expanding part is a mirrored architecture of the contracting part, with max pooling replaced by nearest neighbor upsampling layer. A skip layer that contains a residual module connects each pair of max pooling and upsampling layer so that the spatial information at each resolution will be preserved. Elementwise addition between the skip layer and the upsampled feature map follows to integrate the information across two adjacent resolutions. Both contractive and expanding part utilize large amount of residual modules \cite{he2016deep}. Figure \ref{fig:network_hourglass} (a) shows one HG structure.

One pair of the contractive and expanding network can be viewed as one iteration of prediction. By stacking multiple HG networks together, we can further reevaluate and refine the initial prediction. In our experiment, we find a two-stack network is sufficient to provide satisfactory performance. Adding additional networks only marginally improves the results but at the expense of longer training time. Further, since our stacked HG network is very deep, we also insert auxiliary supervision after each HG network to facilitate the training process. Specifically, we first apply $1\times1$ convolution after each HG to generate an intermediate depth prediction. By comparing the prediction against the ground truth depth, we compute a loss. Finally, the intermediate prediction is remapped to the feature space by applying another $1\times1$ convolution, then added back to the features output from previous HG network. Our two-stack HG network has two intermediate loss, whose weight is equal to the weight of the final loss.

\begin{figure*}[t]
\begin{center}
   \includegraphics[width=1.0\linewidth]{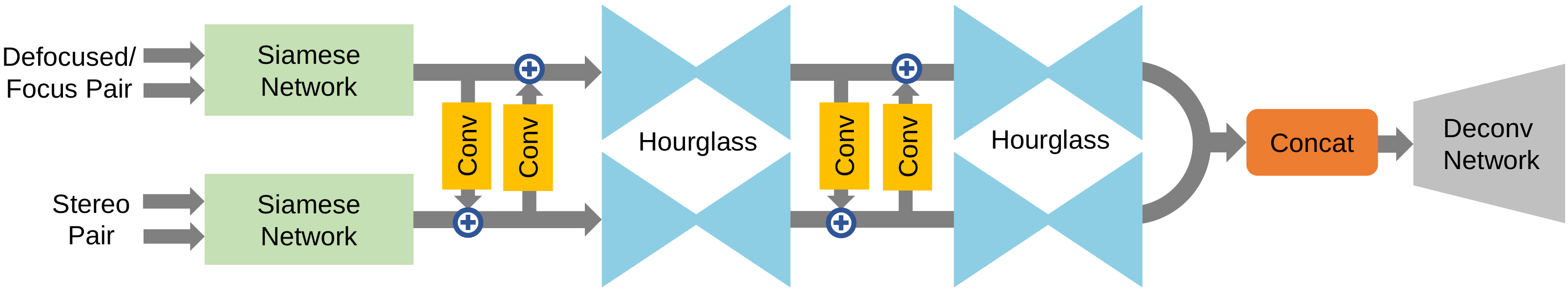}
\end{center}
\vspace{-8pt}
   \caption{Architecture of HG-Fusion-Net. The convolution layers exchange information between networks at various stages, allowing the fusion of defocus and disparity cues.}
\label{fig:hg-fusion-net}
\end{figure*}

Before the two-stack HG network, we add a siamese network, whose two network branches share the same architecture and weights. By using convolution layers that have a stride of 2, The siamese network serves to shrink the size of the feature map, thus reducing the memory usage and computational cost of the HG network. Compared with non-weight-sharing scheme, the siamese network requires fewer parameters and regularizes the network. After the HG network, we apply deconvolution layers to progressively recover the image to its original size. At each scale, the upsampled low resolution features are fused with high resolution features from the siamese network. This upsampling process with multi-scale guidance allows structures to be resolved at both fine- and large-scale. Note that based on our experiment, the downsample/upsample process largely facilitates the training and produces results that are very close to those obtained from full resolution patches. Finally, the network produces pixel-wise disparity prediction at the end. The network architecture is shown in Figure \ref{fig:network_hourglass} (b). For the details of layers, we use $2$-D convolution of size $7\times 7\times 64$ and $5\times 5\times 128$ with stride $2$ for the first two convolution layers in the siamese network. Each residual module contains three convolution layers as shown in Figure \ref{fig:network_hourglass} (c). For the rest of convolution layers, they have kernel size of either $3\times 3$ or $1\times 1$. The input to the first hourglass is of quarter resolution and $256$ channels while the output of the last hourglass is of the same shape. For the Deconv Network, the two $2$-D deconvolution layers are of size  $4\times 4\times 256$ and $4\times 4\times 128$ with stride $2$.

We use one such network for both DfD and stereo, which we call HG-DfD-Net and HG-Stereo-Net. The input of HG-DfD-Net is defocused/focus image pair of the left stereo view and the input of HG-Stereo-Net is stereo pair.

\subsection{Network Fusion}
\label{sec:FusionNet}

The most brute-force approach to integrate DfD and stereo is to directly concatenate the output disparity maps from the two branches and apply more convolutions. However, such an approach does not make use of the features readily presented in the branches and hence neglects cues for deriving the appropriate combination of the predicted maps. Consequently, such naïve approaches tend to average the results of two branches rather than making further improvement, as shown in Table \ref{tab:comparison_networks}.

Instead, we propose HG-Fusion-Net to fuse DfD and stereo, as illustrated in figure \ref{fig:hg-fusion-net}. HG-Fusion-Net consists of two sub-networks, the HG-DfD-Net and HG-Stereo-Net. The inputs of HG-Fusion-Net are stereo pair plus the defocused image of the left stereo view, where the focused image of the left stereo view is fed into both the DfD and stereo sub-network. We set up extra connections between the two sub-networks. Each connection applies a $1\times1$ convolution on the features of one sub-network and adds to the other sub-network. In doing so, the two sub-networks can exchange information at various stages, which is critical for different cues to interact with each other. The $1\times1$ convolution kernel serves as a transformation of feature space, consolidating new cues into the other branch.

In our network, we set up pairs of interconnections at two spots, one at the beginning of each hourglass. At the cost of only four $1\times1$ convolutions, the interconnections largely proliferate the paths of the network. The HG-Fusion-Net can be regarded as an ensemble of original HG networks with different lengths that enables much stronger representation power. In addition, the fused network avoids solving the whole problem all at once, but first collaboratively solves the stereo and DfD sub-problems, then merges into one coherent solution.

In addition to the above proposal, we also explore multiple variants of the HG-Fusion-Net. With no interconnection, the HG-Fusion-Net simply degrades to the brute-force approach. A compromise between our HG-Fusion-Net and the brute-force approach would be using only one pair of interconnections. We choose to keep the first pair, the one before the first hourglass, since it would enable the network to exchange information early. Apart from the number of interconnections, we also investigate the identity interconnections, which directly adds features to the other branch without going through $1\times1$ convolution. We present the quantitative results of all the models in Table \ref{tab:comparison_networks}.

\begin{figure*}[t]
\begin{center}
   \includegraphics[width=1.0\linewidth]{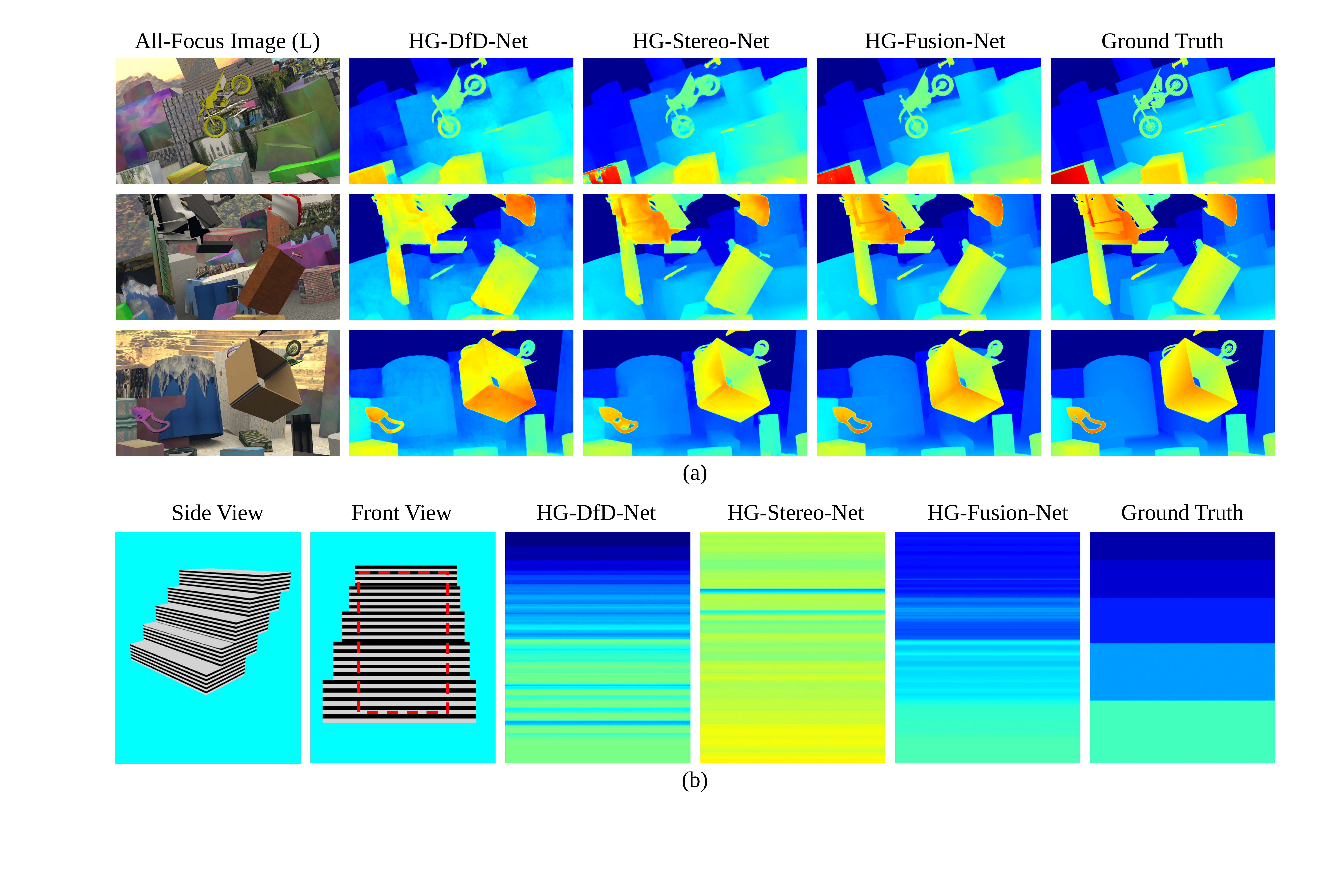}
\end{center}
\vspace{-8pt}
   \caption{Results of HG-DfD-Net, HG-Stereo-Net and HG-Fusion-Net on (a) our dataset (b) staircase scene textured with horizontal stripes. HG-Fusion-Net produces smooth depth at flat regions while maintaining sharp depth boundaries. Best viewed in the electronic version by zooming in.}
\label{fig:result_synthetic}
\end{figure*}

\section{Implementation}

\noindent\textbf{Optimization}
All networks are trained in an end-to-end fashion. For the loss function, we use the mean absolute error (MAE) between ground truth disparity map and predicted disparity maps along with $l_2$-norm regularization on parameters. We adopt MXNET \cite{chen15} deep learning framework to implement and train our models. Our implementation applies batch normalization \cite{ioffe15} after each convolution layer, and use PReLU layer \cite{he15} to add nonlinearity to the network while avoiding the ``dead ReLU''. We also use the technique from \cite{he15} to initialize the weights. For the network solver we choose the Adam optimizer \cite{kingma15} and set the initial learning rate = 0.001, weight decay = 0.002, $\beta1$ = 0.9, $\beta2$ = 0.999. We train and test all the models on a NVIDIA Tesla K80 graphic card.


\noindent\textbf{Data Preparation and Augmentation}
To prepare the data, we first stack the stereo/defocus pair along the channel's direction, then extract patches from the stacked image with a stride of 64 to increase the number of training samples. Recall that the HG network contains multiple max pooling layers for downsampling, the patch needs to be cropped to the nearest number that is multiple of 64 for both height and width. In the training phase, we use patches of size $512\times256$ as input. The large patch contains enough contextual information to recover depth from both defocus and stereo. To increase the generalization of the network, we also augment the data by flipping the patches horizontally and vertically. We perform the data augmentation on the fly at almost no additional cost.

\section{Experiments}
\label{sec:experiments}

\begin{table*}[t]
\begin{center}
 \begin{tabular}{c|c|c|c|c|c}

    \hline
    \ & $>$ 1 px & $>$ 3 px & $>$ 5 px & MAE (px) & Time (s) \\ [0.5ex]
    \hline
    \emph{HG-DfD-Net} & 70.07\% & 38.60\% & 20.38\% & 3.26 & \ 0.24 \\
    \emph{HG-Stereo-Net} & 28.10\% & 6.12\% & 2.91\% & 1.05 & \ 0.24 \\
    \emph{\textbf{HG-Fusion-Net}} & 20.79\% & 5.50\% & \textbf{2.54}\% & \textbf{0.87} & \ 0.383 \\
    \hline
    \hline
    \emph{No Interconnection} & 45.46\% & 10.89\% & 5.08\% & 1.57 & 0.379 \\
    \emph{Less Interconnection} & 21.85\% & \textbf{5.23}\% & 2.55\% & 0.91 & 0.382 \\
    \emph{Identity Interconnection} & 21.37\% & 6.00\% & 2.96\% & 0.94 & 0.382 \\
    \hline
    \hline
    \emph{MC-CNN-fast} \cite{zbontar2016stereo} & 15.38\% & 10.84\% & 9.25\% & 2.76 & \ 1.94 \\
    \emph{MC-CNN-acrt} \cite{zbontar2016stereo} & \textbf{13.95}\% & 9.53\% & 8.11\% & 2.35 & \ 59.77 \\
    \hline

\end{tabular}
\caption{Quantitative results on synthetic data. Upper part compares results from different input combinations: defocus pair, stereo pair and stereo pair + defocused image. Middle part compares various fusion scheme, mainly differentiating by the number and type of interconnection: \emph{No interconnection} is the brute-force approach that only concatenates feature maps after the HG network, before the deconvolution layers. \emph{Less Interconnection} only uses one interconnection before the first hourglass; \emph{Identity Interconnection} directly adds features to the other branch, without applying the $1\times1$ convolution. Lower part shows results of \cite{zbontar2016stereo}. The metrics > 1 px, > 3 px, > 5 px represent the percentage of pixels whose absolute disparity error is larger than 1, 3, 5 pixels respectively. MAE measures the mean absolute error of disparity map.}
\label{tab:comparison_networks}
\end{center}
\end{table*}

\subsection{Synthetic Data}
\label{sec:syntheticExperiments}

We train HG-DfD-Net, HG-Stereo-Net and HG-Fusion-Net separately, and then conduct experiments on test samples from synthetic data. Figure \ref{fig:result_synthetic}(a) compares the results of these three networks. We observe that results from HG-DfD-Net show clearer depth edge, but also exhibit noise on flat regions. On the contrary, HG-Stereo-Net provides smooth depth. However, there is depth bleeding across boundaries, especially when there are holes, such as the tire of the motorcycle on the first row. We suspect that the depth bleeding is due to occlusion, by which the DfD is less affected. Finally, HG-Fusion-Net finds the optimal combination of the two, producing smooth depth while keeping sharp depth boundaries. Table \ref{tab:comparison_networks} also quantitatively compares the performance of different models on our synthetic dataset. Results from Table \ref{tab:comparison_networks} confirm that HG-Fusion-Net achieves the best result for almost all metrics, with notable margin ahead of using stereo or defocus cues alone. The brute-force fusion approach with no interconnection only averages results from HG-DfD-Net and HG-Stereo-Net, thus is even worse than HG-Stereo-Net alone. The network with fewer or identity interconnection performs slightly worse than HG-Fusion-Net, but still a lot better than the network with no interconnection. This demonstrates that interconnections can efficiently broadcast information across branches and largely facilitate mutual optimization. We also compare our models with the two stereo matching approaches from \cite{zbontar2016stereo} in Table \ref{tab:comparison_networks}. These approaches utilize CNN to compute the matching costs and use them to carry out cost aggregation and semi-global matching, followed by post-processing steps. While the two approaches have fewer pixels with error larger than 1 pixel, they yield more large-error pixels and thus have worse overall performance. In addition, their running time is much longer than our models.

We also conduct another experiment on a scene with a staircase textured by horizontal stripes, as illustrated in figure \ref{fig:result_synthetic}(b). The scene is rendered from the front view, making it extremely challenging for stereo since all the edges are parallel to the epipolar line. On the contrary, DfD will be able to extract the depth due to its 2D aperture. Figure \ref{fig:result_synthetic}(b) shows the resultant depths enclosed in the red box of the front view, proving the effectiveness of our learning-based DfD on such difficult scene. Note that the inferred depth is not perfect. This is mainly due to the fact that our training data lacks objects with stripe texture. We can improve the result by adding similar textures to the training set.

\begin{figure*}[h]
\begin{center}
   \includegraphics[width=1.0\linewidth]{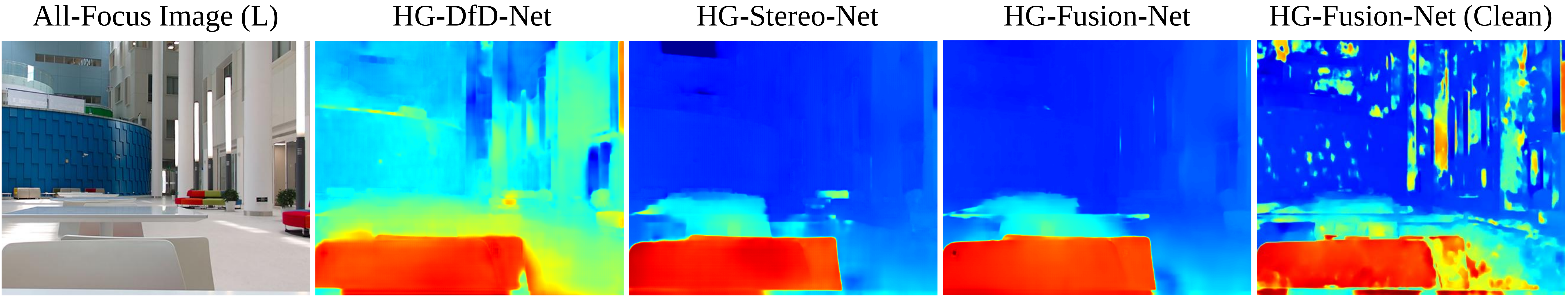}
\end{center}
\vspace{-8pt}
   \caption{Comparisons of real scene results from HG-DfD-Net, HG-Stereo-Net and HG-Fusion-Net. The last column shows the results from HG-Fusion-Net trained by the clean dataset without Poisson noise. Best viewed in color.}
\label{fig:result_realScene1}
\end{figure*}

\begin{figure*}[h]
\begin{center}
   \includegraphics[width=1.0\linewidth]{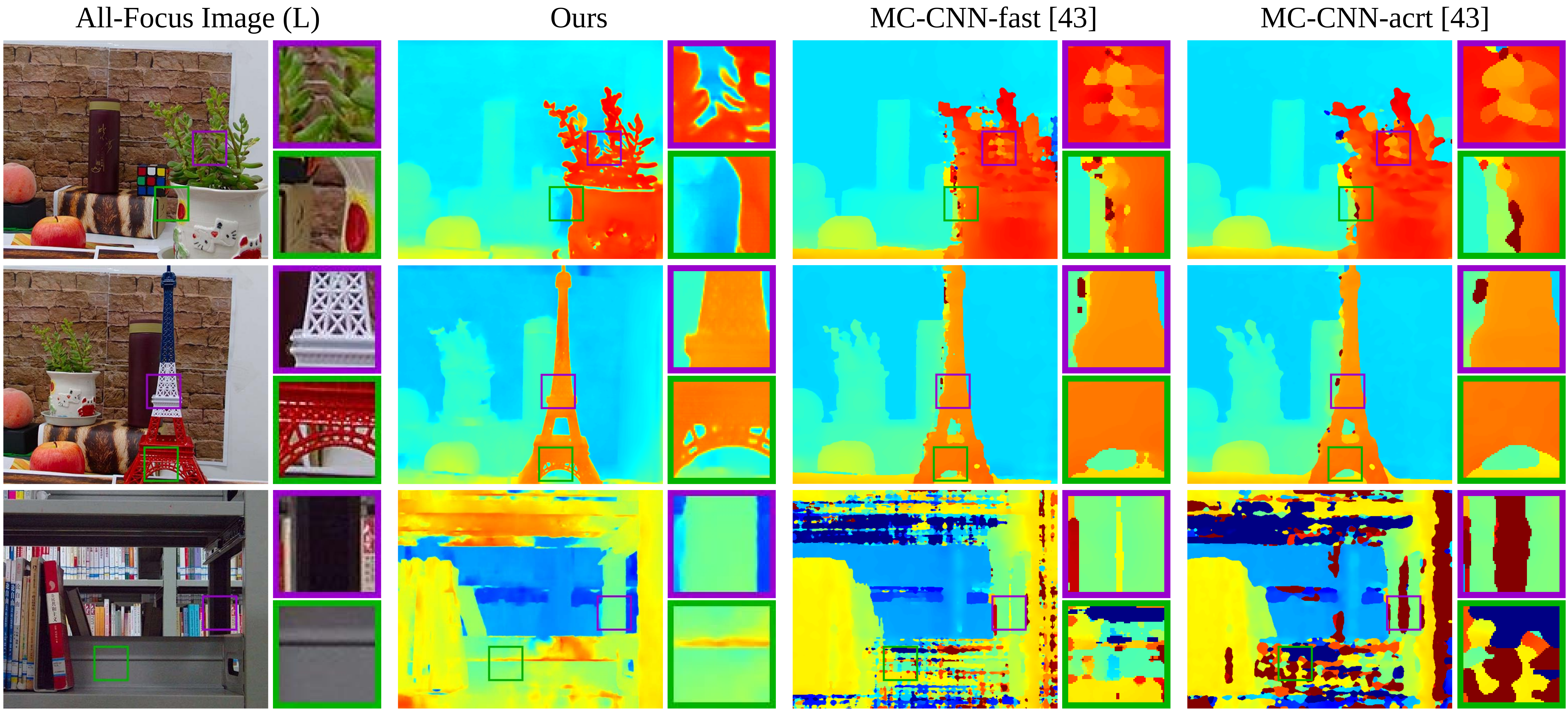}
\end{center}
\vspace{-8pt}
   \caption{Comparisons of real scene results with \cite{zbontar2016stereo}. Best viewed in color.}
\label{fig:result_realScene2}
\end{figure*}

\subsection{Real Scene}
\label{sec:realExperiments}

To conduct experiments on the real scene, we use light field (LF) camera to capture the LF and generate the defocused image. While it is possible to generate the defocused image using conventional cameras by changing the aperture size, we find it difficult to accurately match the light throughput, resulting in different brightness between the all-focused and defocused image. The results using conventional cameras are included in supplementary material.

LF camera captures a rich set of rays to describe the visual appearance of the scene. In free space, LF is commonly represented by two-plane parameterizations $L(u,v,s,t)$, where $st$ is the camera plane and $uv$ is the image plane \cite{levoy96}. To conduct digital refocusing, we can move the synthetic image plane that leads to the following photography equation \cite{ng05}:

\begin{equation}
\label{eqn:refocusing}
    E(s,t) = \iint{L(u,v,u + \frac{s-u}{\alpha}, v+\frac{t-v}{\alpha})}dudv
\end{equation}

By varying $\alpha$, we can refocus the image at different depth. Note that by fixing $st$, we obtain the sub-aperture image $L_{(s^\star t^\star)}(u,v)$ that is amount to the image captured using a sub-region of the main lens aperture. Therefore, Eqn. \ref{eqn:refocusing} corresponds to shift-and-add the sub-aperture images \cite{ng05}.

In our experiment, we use Lytro Illum camera as our capturing device. We first mount the camera on a translation stage and move the LF camera horizontally to capture two LFs. Then we extract the sub-aperture images from each LF using Light Field Toolbox \cite{dansereau13}. The two central sub-aperture images are used to form a stereo pair. We also use the central sub-aperture image in the left view as the all-focused image due to its small aperture size. Finally, we apply the shift-and-add algorithm to generate the defocused image. Both the defocused and sub-aperture images have the size of $625\times433$.

We have conducted tests on both outdoor and indoor scenes. In Fig.\ref{fig:result_realScene1}, we compare the performance of our different models. In general, both HG-DfD-Net and HG-Stereo-Net preserve depth edges well, but results from HG-DfD-Net are noisier. In addition, the result of HG-DfD-Net is inaccurate at positions distant from camera because defocus blurs vary little at large depth. HG-Fusion-Net produces the best results with smooth depth and sharp depth boundaries. We have also trained HG-Fusion-Net on a clean dataset without Poisson noise, and show the results in the last column of Fig.\ref{fig:result_realScene1}. The inferred depths exhibit severe noise pattern on real data, confirming the necessity to add noise to the dataset for simulating real images.

In Fig.\ref{fig:result_realScene2}, we compare our approach with stereo matching methods from \cite{zbontar2016stereo}. The plant and the toy tower in the first two rows present challenges for stereo matching due to the heavy occlusion. By incorporating both DfD and stereo, our approach manages to recover the fine structure of leaves and segments as shown in the zoomed regions while methods from \cite{zbontar2016stereo} either over-smooth or wrongly predict at these positions. The third row further demonstrates the efficacy of our approach on texture-less or striped regions.

\section{Conclusion}

We have presented a learning based solution for a hybrid DfD and stereo depth sensing scheme. We have adopted the hourglass network architecture to separately extract depth from defocus and stereo. We have then studied and explored multiple neural network architectures for linking both networks to improve depth inference. Comprehensive experiments show that our proposed approach preserves the strength of DfD and stereo while effectively suppressing their weaknesses. In addition, we have created a large synthetic dataset for our setup that includes image triplets of a stereo pair and a defocused image along with the corresponding ground truth disparity.

Our immediate future work is to explore different DfD inputs and their interaction with stereo. For instance, instead of using a single defocused image, we can vary the aperture size to produce a stack of images where objects at the same depth exhibit different blur profiles. Learning-based approaches can be directly applied to the profile for depth inference or can be combined with our current framework for conducting hybrid depth inference. We have presented one DfD-Stereo setup. Another minimal design was shown in \cite{takeda13}, where a stereo pair with different focus distance is used as input. In the future, we will study the cons and pros of different hybrid DfD-stereo setups and tailor suitable learning-based solutions for fully exploiting the advantages of such setups.

\section{Acknowledgments}

This project was supported by the National Science Foundation under grant CBET-1706130.

{\small
\bibliographystyle{ieee}
\bibliography{egbib}
}

\onecolumn
\newpage
\appendix
\section{Results On Data Captured By Conventional Cameras}
\begin{figure}[H]
\begin{center}
   \includegraphics[width=0.99\linewidth]{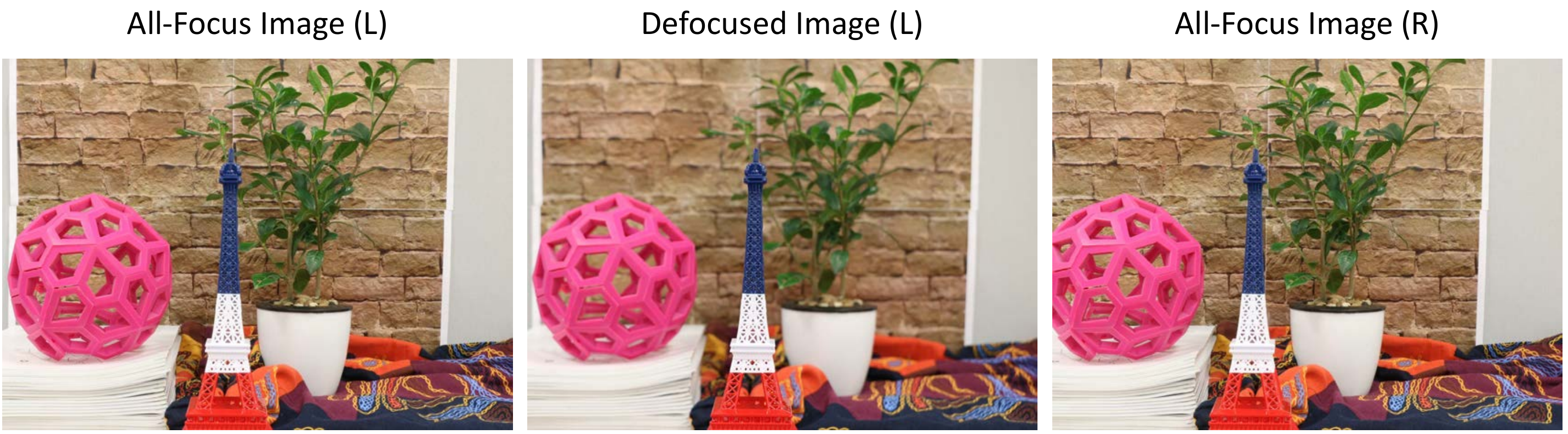}
\end{center}
\vspace{-8pt}
   \caption{An Example of our captured data.}
\label{fig:supple0}
\end{figure}

\begin{figure}[H]
\begin{center}
   \includegraphics[width=0.99\linewidth]{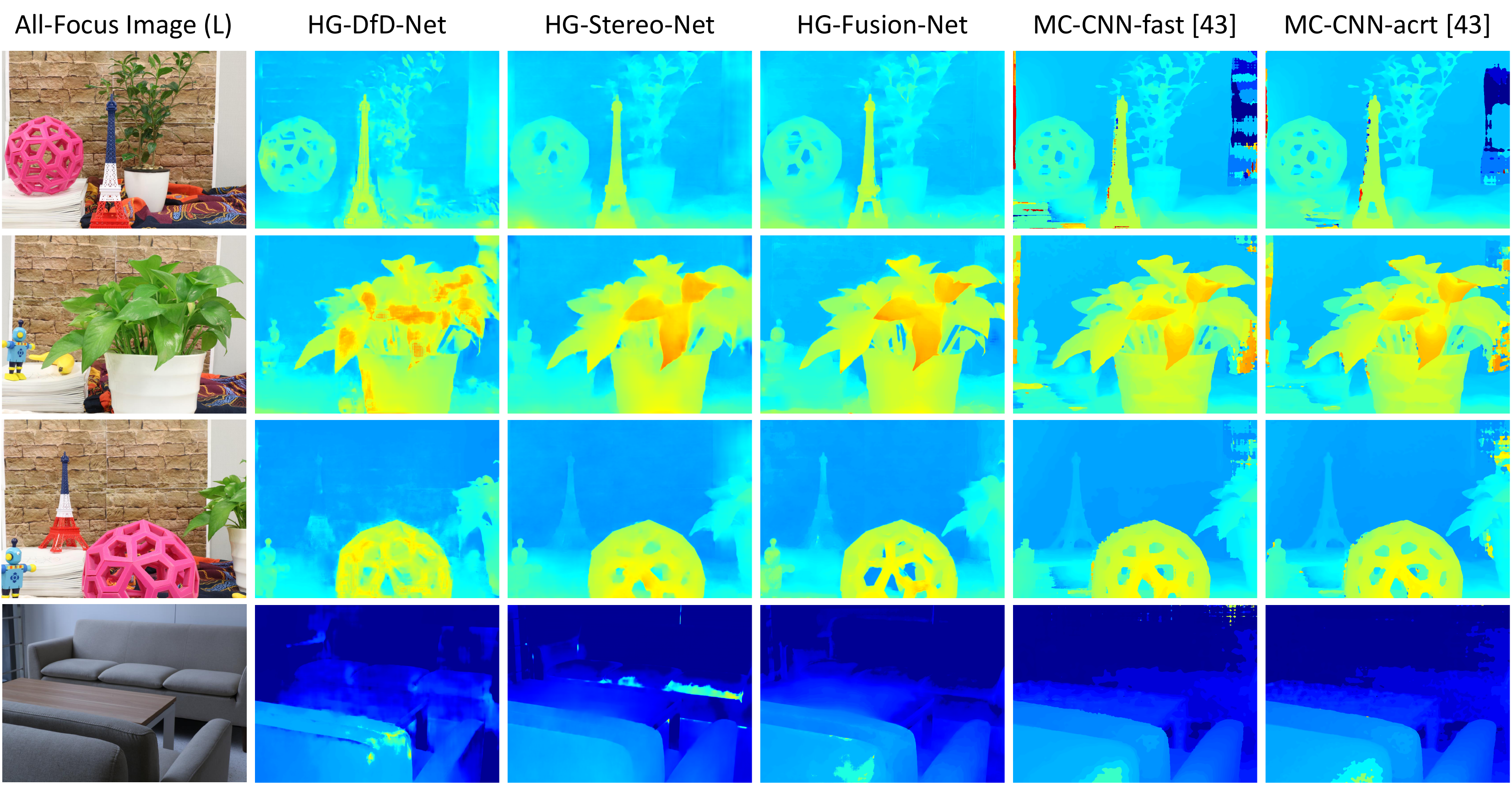}
\end{center}
\vspace{-8pt}
   \caption{Results on data captured by \emph{DSLR} cameras. The first three columns are results of our different models while the last two columns are results of \cite{zbontar2016stereo}.}
\label{fig:supple1}
\end{figure}

We have captured real scene image triplets using a pair of \emph{DSLR} cameras. We capture the defocused left image by changing the aperture of cameras.
Figure \ref{fig:supple0} shows one example of image triplet. Notice that the brightness of all-focus and defocused left image is inconsistent near image borders.
We also compare our results with \cite{zbontar2016stereo} on our data captured with \emph{DSLR} cameras, as shown in Figure \ref{fig:supple1}.

\end{document}